\documentclass[11pt]{article}

\usepackage[preprint]{src/acl}

\usepackage{times}
\usepackage{latexsym}
\usepackage{multirow}

\usepackage[T1]{fontenc}

\usepackage[utf8]{inputenc}

\usepackage{microtype}

\usepackage{inconsolata}

\usepackage{mathtools}
\usepackage{enumitem}
\usepackage{graphicx}
\usepackage{amsmath, amssymb}
\usepackage{algorithm}
\usepackage{algpseudocode}
\usepackage{booktabs}
\usepackage{todonotes}
\title{Structured-Sparse Attention for Entity Tracking with Subquadratic Sequence Complexity}

\author{
Hangyue Zhao\thanks{Equal contribution.}\thanks{Correspondence to \texttt{hangyue.zhao@dauphine.eu}}\textsuperscript{1}
\and
Paul Caillon\footnotemark[1]\textsuperscript{2}
\and
Erwan Fagnou\footnotemark[1]\textsuperscript{2}
\and
Alexandre Allauzen\textsuperscript{1,2}
\\
\textsuperscript{1} ESPCI PSL, Paris, France \\
\textsuperscript{2} LAMSADE, Université Paris Dauphine - PSL, Paris, France
}

\begin{document}
\maketitle
\begin{abstract}
Entity tracking requires maintaining and updating latent states for entities and attributes over long sequences. Recent task-specific attention operators can compress deep Transformer stacks into a few layers by performing multi-hop state propagation within a single layer, but their dense evaluation remains expensive. We show that in this setting, learned attention is strongly structured: most mass concentrates in local block-diagonal neighborhoods with a light cross-block residue. Exploiting this, we derive a blockwise evaluation of a resolvent-style operator that keeps within-block interactions exact and routes cross-block interactions through a reduced system.
The resulting evaluation is subquadratic in sequence length $O(n^{4/3}d)$ (and $O(n^{7/3})$ when $d\approx n$).
On controlled tracking benchmarks, our method matches the dense operator’s accuracy while reducing wall-clock time by $12-29\%$ under a standardized measurement protocol, and is up to $2.4 \times$ faster than a compact dense Transformer at comparable exact-match accuracy. We further provide ablations over block size and model capacity, and identify a limitation: performance collapses when the number of simultaneously evolving properties exceeds the number of attention heads.
\end{abstract}

\section{Introduction}

Maintaining a consistent latent state for entities and attributes across long sequences, also known as \emph{entity tracking}, is critical for document-level information extraction, dialogue state tracking, and long-context QA/narrative understanding, requiring models to integrate dispersed mentions, resolve coreference, and update attributes as events unfold \citep{zheng-etal-2024-comprehensive,budzianowski2018multiwoz,dasigi-etal-2019-quoref,dalvi2018tracking,lee2017end,henaff2017tracking}.

A common recipe is to stack Transformer layers and rely on learned attention to implement routing and memory \citep{NIPS2017_3f5ee243}. In its standard form, attention computes a dense interaction matrix over all token pairs,
\(A=\mathrm{Softmax}\!\big(\frac{QK^\top}{\sqrt{d_k}}+M\big)\), and outputs \(Y=AV\),
which is expressive but incurs at least quadratic cost in sequence length \(n\).
This cost motivates a large body of efficient/long-context variants based on locality, sparsity, low-rank structure, or kernellization \citep{kitaev2020reformer,beltagy2020longformer,zaheer2020bigbird,katharopoulos2020linear,wang2020linformer,choromanski2021performer,child2019sparse}, and standardized benchmarks emphasize reasoning over long inputs \citep{tay2021long,shaham2022scrolls}. Complementary work explores structure-aware or task-specific operators for entity-centric reasoning \citep{kim2017structured,gerritse2022entity,Fagnou_2024} as well as systems-level accelerations that improve IO/parallelism without changing the operator \citep{dao2022flashattentionfastmemoryefficientexact}.
From a depth perspective, standard stacked attention must grow with the number of state changes \(K\) to propagate information: \(\mathcal{O}(\log K)\) layers suffice (and can be necessary) for \(K\) updates, motivating depth-compressing operators such as ChaCAL~\citep{Fagnou_2024}. A more detailed discussion of related work is provided in Appendix~\ref{app:extended_related}. 
\newline
We focus on the entity-tracking regime and make a simple observation: learned attention maps are \emph{structured and sparse}. Most mass concentrates within local, block-diagonal neighborhoods around “state-carrying” tokens with a light cross-block residue (matching analyses of attention selectivity \citep{clark2019entity}). Exploiting this regularity, we introduce a \emph{blockwise inverse formulation} that keeps within-block interactions exact while routing off-block mass through a compressed residual path, yielding a fast evaluation with \textbf{subquadratic sequence complexity}:
\[
T(n,d)=\tilde{\mathcal{O}}(n^{4/3}d),
\]
and \(\tilde{\mathcal{O}}(n^{7/3})\) when width scales with length (e.g., \(d\!\approx\!n\)).\footnote{As standard, we express complexity in $n$, treating $d$ as independent, and also report the coupled regime $d \approx n$; polylogarithmic factors are absorbed into $\tilde{\mathcal O}$.}
This matters for long-context modeling and composes with low-level kernels \citep{dao2022flashattentionfastmemoryefficientexact}.

Empirically, our operator matches a dense task specific attention layer in accuracy while reducing wall-clock time by $12 - 29\%$ under a standardized protocol and achieves up to $2.4\times$ faster inference
than a compact Transformer at comparable exact-match accuracy. Our contributions are: (i) an empirical, structured-sparse view of attention in entity tracking; (ii) a blockwise inverse formulation with subquadratic sequence complexity; (iii) practical speedups at matched accuracy with ablations over block size, sequence length, and head count; and (iv) a characterization of a capacity limit whereby single-head attention fails as the number of simultaneously evolving properties exceeds the number of heads. We detail the method and analysis next, then present experiments and discuss limitations.

\begin{figure*}[t]
  \centering
  \includegraphics[width=0.926\linewidth]{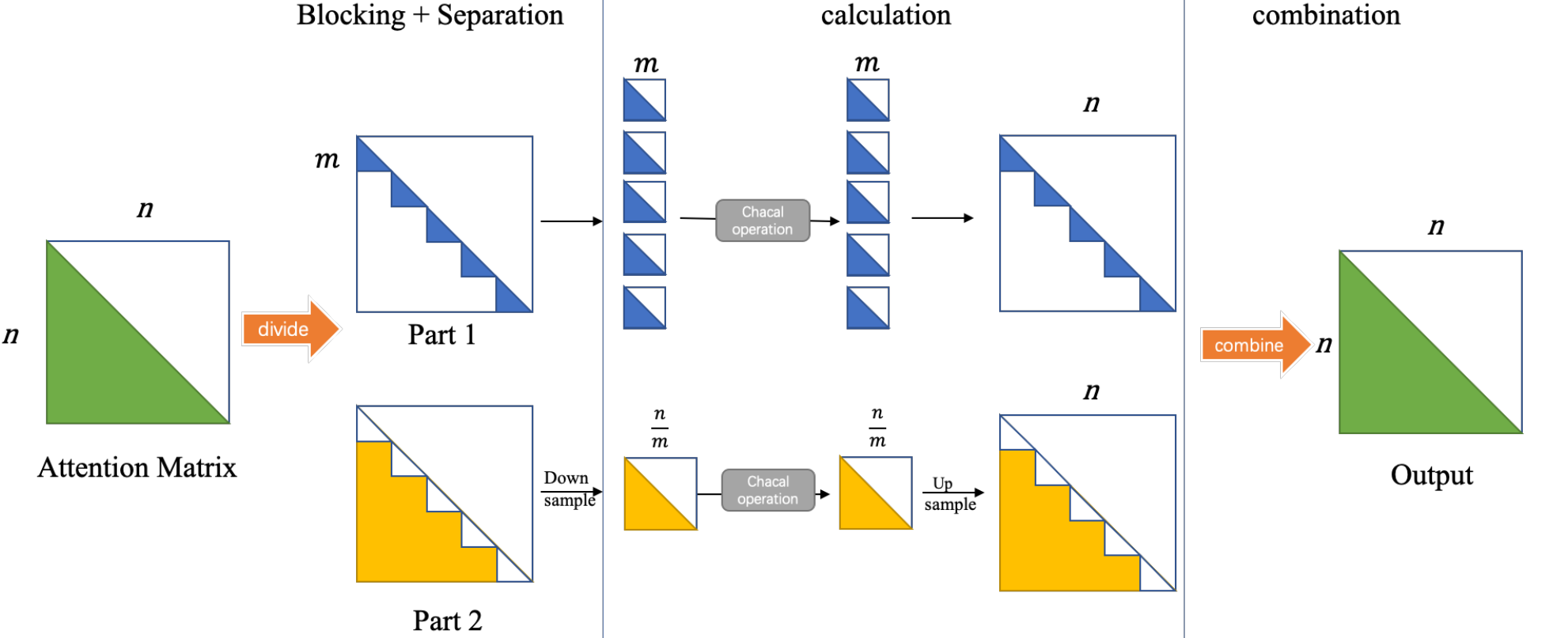}
  \caption{\textbf{Blockwise evaluation of resolvent attention.}
Split the causal attention matrix $A\in\mathbb{R}^{n\times n}$ into $k=n//m$ contiguous $m\times m$ tiles.
\emph{Local branch:} apply $\mathcal{S}_\gamma(A)=(1-\gamma)A(I-\gamma A)^{-1}$ exactly on the block diagonal.
\emph{Residual branch:} compress off-block interactions to a $k\times k$ system (down-sample), apply the same operator, then lift back (up-sample).
The final output is the sum of both branches.}

  \label{fig:blockdecomp}
\end{figure*}

\section{Method: Blockwise Efficient Attention for Entity Tracking}
\label{sec:method}
In the following section, we present a blockwise evaluation scheme for resolvent-style attention. The construction is motivated by the sparsity patterns analyzed later in Section~\ref{sec:sparsity-memory}, together with the observation that, in entity-tracking settings, attention often exhibits strong local block structure. Given a causal attention matrix $A$ and values $V$, we therefore approximate the dense operator by (i) decomposing $A$ into a block-diagonal term, which captures dominant local interactions, and an off-block residual, which captures the remaining cross-block interactions; (ii) applying the resolvent exactly on each diagonal tile; and (iii) evaluating the residual through a reduced $k \times k$ system, with $k = n/m$, before lifting it back to the original resolution. The final output is obtained by summing the local and residual branches (Figure~\ref{fig:blockdecomp}).


\subsection{Resolvent operator (ChaCAL)}
We build on Chain and Causal Attention (ChaCAL) \citep{Fagnou_2024}, which replaces the standard one-hop map $AV$
by a resolvent-style operator that aggregates multi-hop propagation in a single layer.
For masked single-head attention $A\in\mathbb{R}^{n\times n}$ and values $V\in\mathbb{R}^{n\times d}$, define
\begin{equation}
\begin{aligned}
\mathcal{S}_\gamma(A) \;&\coloneqq\; (1-\gamma)A\,(I-\gamma A)^{-1}, \qquad \gamma\in[0,1),
\\
Y \;&=\; \mathcal{S}_\gamma(A)\,V.
\end{aligned}
\label{eq:resolvent}
\end{equation}
With causal masking, $A$ is lower triangular, hence so is $(I-\gamma A)$, with diagonal entries
$1-\gamma A_{ii}\ge 1-\gamma>0$; therefore $(I-\gamma A)^{-1}$ can be applied via forward substitution
\textbf{Multi-hop interpretation.}
Using the Neumann series (valid for $\gamma\in[0,1)$ and row-stochastic $A$),
\begin{equation}
\begin{aligned}
(I-\gamma A)^{-1} \;&=\; \sum_{t\ge0} \gamma^t A^t,
\\
\mathcal{S}_\gamma(A) \;&=\; (1-\gamma)\sum_{t\ge1}\gamma^{t-1} A^t,
\end{aligned}
\end{equation}
so $\mathcal{S}_\gamma$ aggregates $A, A^2, A^3,\ldots$ (multi-hop propagation) in a single layer.
As detailed in~\cite{Fagnou_2024}, in entity-tracking settings, each multiplication by $A$ propagates information by one hop along the causal attention graph. The resolvent thus compresses multi-hop state propagation into a single layer, emphasizing shorter, higher-probability paths while geometrically down-weighting longer ones by $\gamma^{t-1}$. This is well suited to tasks such as BOXES, where prediction depends on composing successive state updates.

\textbf{Block decomposition}
Partition the sequence into $k=n/m$ contiguous blocks of size $m$ (assume $m\mid n$ for clarity; see edge cases below):
\begin{equation}
\begin{aligned}
A \;&=\; A_{\mathrm{blk}} + A_{\mathrm{res}},\\
A_{\mathrm{blk}} &= \mathrm{BlockDiag}(A_1,\dots,A_k).
\label{eq:block-split}
\end{aligned}
\end{equation}
We evaluate the two branches and sum:
\begin{equation}
\begin{aligned}
Y \;&=\; Y_{\mathrm{blk}} \;+\; Y_{\mathrm{res}}.
\end{aligned}
\label{eq:sum}
\end{equation}

\subsection{Local (exact) branch}
Apply the \emph{same} operator to each diagonal tile and concatenate:
\begin{equation}
\begin{aligned}
Y_{\mathrm{blk}}^{(i)} \;&=\; \mathcal{S}_\gamma(A_i)\,V_i
 \quad i=1,\dots,k,
\\
Y_{\mathrm{blk}} \;&=\; \mathrm{concat}\!\left(Y_{\mathrm{blk}}^{(1)},\dots,Y_{\mathrm{blk}}^{(k)}\right).
\end{aligned}
\label{eq:local}
\end{equation}
Because each $A_i$ is lower triangular (causal within block), $(I-\gamma A_i)^{-1}$ is evaluated by a forward-substitution triangular solve in $\mathcal{O}(m^2 d)$ and preserves the exact masked semantics on the tile.

\subsection{Residual (reduced) branch}
We evaluate off-block interactions in a compressed space. Let $P\!\in\!\mathbb{R}^{k\times n}$ be a fixed row down-sampler that maps each contiguous group of $m$ rows to one row (e.g., average, strided pick; any fixed rule suffices). Let $U\!\in\!\mathbb{R}^{n\times k}$ be the corresponding up-sampler (e.g., nearest-neighbor replication). Define
\begin{equation}
\begin{aligned}
\widetilde{A}_{\mathrm{res}}=PA_{\mathrm{res}}P^\top \in \mathbb{R}^{k\times k},
\;
\widetilde{V} = P\,V \in \mathbb{R}^{k\times d}.
\end{aligned}
\label{eq:reduced}
\end{equation}
Evaluate the same operator at reduced size and lift back:
\begin{equation}
\widetilde{Y}_{\mathrm{res}} \,=\, \mathcal{S}_\gamma(\widetilde{A}_{\mathrm{res}})\,\widetilde{V}
\qquad
Y_{\mathrm{res}} =\; U\,\widetilde{Y}_{\mathrm{res}}.
\label{eq:residual}
\end{equation}
If $P$ preserves order (contiguous pooling), then $\widetilde{A}_{\mathrm{res}}$ inherits block-level causality (lower triangular); $(I-\gamma \widetilde{A}_{\mathrm{res}})$ is triangular with positive diagonal and admits forward substitution as well.

\subsection{Masking and correctness.}
Our construction preserves the masking semantics of the dense operator at every stage. 
First, the local branch in Eq.~\ref{eq:local} applies the same masked logits as dense attention, but restricted to each tile, so within-block behavior is unchanged. 
Second, cross-block interactions are limited to the residual matrix $A_{\mathrm{res}}$, which contains only off-block entries; the compressed system $P(\cdot)P^\top$ preserves this structure and the lifting operator $U$ injects updates only into the corresponding block rows.
Finally, both branches respect causality: maintaining the triangular structure ensures no future information can be introduced.

\subsection{Complexity analysis}
Let $n=km$ and treat width $d$ as independent of $n$.
The local branch applies $k$ triangular solves on $m\times m$ tiles, costing
$\mathcal{O}(k\,m^2d)=\mathcal{O}(nmd)$.
The residual branch forms and applies the reduced $k\times k$ operator in
$\mathcal{O}(k^2d)=\mathcal{O}((n/m)^2d)$ (down/up-sampling is $\mathcal{O}(nd)$ and negligible).
Thus,
\[
T(n,d;m)=\mathcal{O}(nmd)+\mathcal{O}((n/m)^2d).
\]
Balancing the two terms gives $m^\star=\Theta(n^{2/3})$ and
$T(n,d;m^\star)=\widetilde{\mathcal{O}}(n^{4/3}d)$.
In the coupled regime $d\approx n$, this becomes $\widetilde{\mathcal{O}}(n^{7/3})$. For more details, see appendix~\ref{sec:appendix_complexity_details}.

\subsection{Full algorithm (single head)}
\begin{algorithm}[H]
\caption{Blockwise Efficient Attention}
\label{alg:blockwise}
\begin{algorithmic}[1]
\Require Masked (causal) attention $A\in\mathbb{R}^{n\times n}$, values $V\in\mathbb{R}^{n\times d}$, block size $m$, scalar $\gamma\in[0,1)$; down-sampler $P\in\mathbb{R}^{k\times n}$, up-sampler $U\in\mathbb{R}^{n\times k}$, where $k=n/m$.
\State Partition $A$ into $k$ diagonal tiles $\{A_i\}_{i=1}^k$ of size $m\times m$; partition $V$ accordingly into $\{V_i\}$
\State \textbf{Local (exact):} For each $i$, compute $B_i \leftarrow I-\gamma A_i$ 
, $R_i \leftarrow A_i V_i$; then $Y_{\mathrm{blk}}^{(i)} \leftarrow (1-\gamma)\,\texttt{triSolve}(B_i,\; R_i)$
\State Assemble $Y_{\mathrm{blk}} \leftarrow \mathrm{concat}(Y_{\mathrm{blk}}^{(1)},\dots,Y_{\mathrm{blk}}^{(k)})$
\State\resizebox{0.91\columnwidth}{!}{\textbf{Residual:} $A_{\mathrm{res}} \leftarrow A - \mathrm{BlockDiag}(A_1,\dots,A_k)$}
\State $\widetilde{A}_{\mathrm{res}} \leftarrow P\,A_{\mathrm{res}}\,P^\top$;\quad $\widetilde{V} \leftarrow P\,V$
\State $\widetilde{B} \leftarrow I-\gamma \widetilde{A}_{\mathrm{res}}$ 
, $\widetilde{R} \leftarrow \widetilde{A}_{\mathrm{res}}\,\widetilde{V}$
\State $\widetilde{Y}_{\mathrm{res}} \leftarrow (1-\gamma)\,\texttt{triSolve}(\widetilde{B},\; \widetilde{R})$
\State $Y_{\mathrm{res}} \leftarrow U\,\widetilde{Y}_{\mathrm{res}}$
\State \Return $Y \leftarrow Y_{\mathrm{blk}} + Y_{\mathrm{res}}$
\end{algorithmic}
\end{algorithm}

\begin{table*}[t]
\centering
\setlength{\tabcolsep}{3pt}
\resizebox{\textwidth}{!}{
\begin{tabular}{@{}l c c c c c c c c c c c c c c@{}}
\toprule
& \multicolumn{3}{c}{\textbf{Transformer, L=2}}
& \multicolumn{3}{c}{\textbf{Transformer, L=5}}
& \textbf{ChaCAL, L=2}
& \multicolumn{7}{c}{\textbf{Block-ChaCAL (ours), L=2}} \\
\cmidrule(lr){2-4}\cmidrule(lr){5-7}\cmidrule(lr){9-15}
& Dense & BigBird & Local 
& Dense & BigBird & Local 
& $m{=}n$
& $n//2$ & $n//3$ & $n//4$ & $n//5$ & $n//6$ & $n//7$ & $n//8$ \\
\midrule
Accuracy (\%)
& 97.20 & 94.49 & 89.49 
& 99.87 & 99.37 & 96.97 
& 99.99
& 99.99 & 99.98 & 99.91 & 99.92 & 99.85 & 99.85 & 99.90 \\
EM (\%)
& 58.8 & 44.51 & 30.12 
& 97.0 & 88.24 & 71.67 
& 100.00
& 100.00 & 99.99 & 97.40 & 97.00 & 94.71 & 94.92 & 96.60 \\
Time (s)
& 1.84 & -- & --
& 4.35 & -- & --
& 2.50
& 2.20 & 1.90 & 1.79 & 1.78 & 1.77 & 1.89 & 1.95 \\
\bottomrule
\end{tabular}}
\caption{Results on \textsc{Boxes}. Within each Transformer depth (L=2 and L=5), we report dense attention and two efficient-attention variants (local window of size $w=32$ and BigBird-style sparse attention (with local window of size $w=32$ and number of random tokens $r=3$). ChaCAL is the resolvent operator with block size $m{=}n$; Block-ChaCAL sweeps block sizes. Results are averaged over three random seeds (std.\ below $0.5\%$ for all entries).}
\label{tab:boxes}
\end{table*}

\section{Experiments}
\label{sec:experiments}

We empirically assess the proposed blockwise evaluation of the resolvent-style attention operator on controlled entity-tracking tasks. We compare against (i) a compact Transformer with dense attention, (ii) the single-layer entity-tracking operator (ChaCAL; \citet{Fagnou_2024}), and (iii) representative generic efficient-attention baselines (sliding-window local attention and BigBird-style block-sparse attention). We further analyze sparsity and capacity limits, and evaluate language modeling and code reasoning as natural proxies for entity tracking. 
All experiments use a single NVIDIA A6000 GPU, batch size \(256\), AdamW with learning rate \(3{\times}10^{-4}\), and a maximum of \(25{,}000\) training steps unless stated otherwise.
Timing is reported as average evaluation time in seconds for the full test set.
Unless otherwise stated, masking is causal; evaluation uses the same batch size for all methods.
A detailed description of the experimental setup can be found in Appendix~\ref{app:exp_details}.

\subsection{Tasks and metrics}\label{sec:tasks-metrics}

\textbf{Toy (chains).}
A synthetic tracking task that isolates long-range, multi-hop dependencies.
Sequences have length \(n=128\), built from 16 blocks of size \(8\), where each block (after the first) is a shuffled permutation of indices into the previous block; supervision is the multi-step chained lookup, with longest chains of length 15.
We measure token accuracy and probe sparsity with masking interventions (\emph{Top-$k$}) applied during training/inference.

\textbf{Boxes.}
A natural-language entity-tracking task in the \emph{advanced} setup introduced by \citet{Fagnou_2024}: each instance describes an initial state (e.g., “The camera is in Box F”) followed by a sequence of \textsc{Move} operations, of the form “move the contents of Box H to Box F”, forcing reconstruction of prior contents.

\textbf{Language Modeling (LM).}
We evaluate perplexity (PPL; lower is better) with a fixed tokenizer and context length \(n=1024\).
We report (i) fine-tuning from a GPT-2 model pretrained with dense attention (OpenWebText) and (ii) pretraining from scratch on WikiText-103. 

\textbf{Code reasoning (Copilot).}
We fine-tune models on Copilot under a controlled substitution protocol: in a 12-layer decoder-only backbone, we replace attention in \emph{only layer 2} by the mechanism under study; all other layers remain standard. We report classification accuracy.

\textbf{SCROLLS (long-context sequence to sequence).}
We consider BART-base with input length up to \(n=1024\) and evaluate SCROLLS tasks spanning summarization (ROUGE), question answering (token-level F1), and classification-style settings (EM).
To isolate the effect of the mechanism while minimizing distribution shift, we follow a replacement protocol analogous to the Copilot experiment: we keep the \emph{encoder non-causal} (standard bidirectional self-attention) and substitute the attention mechanism in \emph{only the first decoder self-attention layer}; all other attention modules remain dense.
This design mirrors the intuition that an early decoder routing layer can influence long-range propagation during generation while preserving most of the pretrained architecture.

\subsection{Models}\label{sec:models}

In the following, we denote for each model its hidden size \(d_{\mathrm{model}}\); MLP size \(d_{\mathrm{ff}}\); number of heads \(h\); and number of layers \(L\).
We also denote the sequence length $n$, the used block size $m = n//k$, with $k \in [2, 8]$.

\textbf{Standard Transformer (dense).}
We study causal self-attention with feed-forward blocks as a first baseline. The by-default settings for each task are shared across models unless stated otherwise:
\begin{itemize}[nosep]
    \item \textbf{Toy} uses \(L{=}1, h{=}8, d_{\mathrm{model}}{=}512, d_{\mathrm{ff}}{=}2048\);
    \item \textbf{Boxes} uses \(L{=}2\) (one standard Transformer layer plus one mechanism layer), \(h{=}8, d_{\mathrm{model}}{=}512, d_{\mathrm{ff}}{=}2048\) and serves as wall-clock/accuracy reference;
    \item \textbf{LM/Code} use a GPT-2–style backbone with \(L{=}12, h{=}12, d_{\mathrm{model}}{=}768, d_{\mathrm{ff}}{=}3072\).
    \item \textbf{SCROLLS} uses BART-base with \(L_{\mathrm{enc}}{=}6\), \(L_{\mathrm{dec}}{=}6\), \(h{=}12\), \(d_{\mathrm{model}}{=}768\), \(d_{\mathrm{ff}}{=}3072\), and maximum input length \(n{=}1024\) (decoder outputs up to 256 tokens).
\end{itemize}

\textbf{Replacement protocol (applies to all mechanism variants).}
\textbf{Toy}: replace the sole attention layer. \textbf{Boxes}: replace the mechanism layer only (the other layer remains standard). \textbf{LM}: swap attention in all 12 layers (pre-trained setting). \textbf{Code}: substitute \emph{only} layer 2 (controlled substitution). \textbf{SCROLLS}: substitute \emph{only} the first layer of decoder (controlled substitution)

\textbf{Single-layer operator (ChaCAL).}
We study the resolvent-style attention of \citet{Fagnou_2024}, replacing the map \(AV\) by
\(
C(A,V)=(1-\gamma)\,A\,(I-\gamma A)^{-1}V,
\)
with \(A\) the causal attention matrix. We apply the above replacement protocol.

\textbf{Blockwise evaluation (ours).}
We apply the \emph{blockwise resolvent operator} introduced in Section~\ref{sec:method}: the resolvent is computed \emph{exactly} on block-diagonal \(m{\times}m\) tiles (causal mask preserved within each tile), while cross-tile interactions are handled in a reduced \(k{\times}k\) system with \(k{=}n//m\). We sweep \(m\in\{n//2,n//3,\ldots,n//8\}\); with \(m{=}n\) recovering full ChaCAL. We apply the same replacement protocol.

\textbf{Generic efficient-attention baselines.}
To address comparisons with established efficient attention, we evaluate:
(i) \emph{local sliding-window causal attention} (window size 32) and
(ii) a \emph{BigBird-style} causal sparse pattern with local window 32 plus 3 random links per token.

\subsection{Latency–accuracy trade-offs on \textsc{Boxes}}\label{sec:boxes-main}

Table~\ref{tab:boxes} summarizes accuracy, exact match (EM), and average evaluation time across block sizes, along with generic efficient-attention baselines.

\textbf{Perfect-EM parity.}
With \(m\in\{n//2,n//3\}\), the blockwise layer matches the single-layer operator’s quality (Acc/EM \(\approx\) 100\%/99.99\%) while reducing ChaCAL's latency from \(2.50\)s to \(2.20\)s and \(1.90\)s (\(\mathbf{12\%}\) and \(\mathbf{24\%}\) faster).

\textbf{Transformer-level performance.}
For \(m\in\{n//4,n//5\}\), EM remains \(\approx 97\%\), on par with the 5-layer dense Transformer (97.0\%). However, runtime drops from \(4.35\)s to \(1.79\)s and \(1.78\)s (\(\mathbf{2.43\times}\) and \(\mathbf{2.45\times}\) faster), placing blockwise on the Pareto frontier: same EM as a much deeper stack at well under half the latency.

\textbf{Fast mid-accuracy operating points.}
Smaller blocks \(m\in\{n//6,n//7,n//8\}\) yield EM \(=\) 94.71–96.60\% with \(1.77\)–\(1.95\)s. The best trade-off occurs at \(m{=}n//6\) (94.71\% EM at \(\mathbf{1.77}\)s; \(\mathbf{1.97\times}\) faster than a 4-layer Transformer at similar EM). While \(n//7\) and \(n//8\) remain faster than the 4-layer baseline, they become slightly slower than \(n//6\), consistent with the cost decomposition: the exact local path scales with \(m\) whereas the reduced path scales with \((n/m)^2\); finer tiling also weakens memory locality, pushing total time upward once past the empirical optimum.

\textbf{Efficient-attention baselines and the depth theorem connection.}
Local and BigBird-style sparse attention underperform sharply at small depth on \textsc{Boxes}: at 2 layers they reach only 30.12\% EM (local) and 44.51\% EM (BigBird), and even at 5 layers they reach only 71.67\% and 88.24\% EM, respectively (Table~\ref{tab:boxes}). 
This empirical pattern matches the ChaCAL depth-theorem perspective: standard attention layers (dense or sparse) propagate information only one hop per layer in the underlying state-update graph, yielding a lower bound of \(\Omega(\log(\#\text{updates}))\) layers for tasks such as \textsc{Boxes} (up to 31 updates). In contrast, resolvent-style operators explicitly implement multi-hop propagation within one layer via \((I-\gamma A)^{-1}\), which is why ChaCAL and blockwise variants attain near-perfect EM at shallow depth. 

\textbf{Summary.}
Table~\ref{tab:boxes} yields three takeaways: (i) at near-perfect EM, blockwise matches quality with \textbf{12–24\%} lower time than ChaCAL; (ii) at EM \(\approx\) 97\%, it meets a 5-layer dense Transformer at \(\sim\)\textbf{2.4\(\times\)} lower latency; (iii) for EM 95–97\%, \(m{=}n//6\) offers the best speed/quality trade-off.

\subsection{Ablation study on \textsc{Boxes}}

We ablate the proposed layer at block size \(m{=}n/3\) by replacing only the 2nd encoder layer (others remain standard) and isolating its two branches (Section~\ref{sec:method}). Table~\ref{tab:ablation_branches} shows: \emph{local-only} already attains strong EM (96.68\%), \emph{residual-only} underperforms (6.52\% EM), and \emph{local+residual} recovers near-perfect scores (99.99\%/99.99\%). Thus, the local branch carries most of the tracking signal, while the residual branch complements it; the combination is required for peak performance.

The large gap between \emph{residual-only} and \emph{local-only} shows that precise one-hop routing within tiles is the key prerequisite; without it, the reduced cross-tile path cannot reliably assemble multi-hop flows. Adding the residual branch atop a strong local path fixes remaining cross-block inconsistencies, recovering the last EM points. The low EM but moderate token accuracy for \emph{residual-only} further indicates scattered errors that break sequence-level correctness.

\begin{table}[ht]
\centering
\small
\setlength{\tabcolsep}{5pt}
\begin{tabular}{@{}l rr@{}}
\toprule
\textbf{Ablation setting} & \textbf{EM (\%)} & \textbf{Accuracy (\%)} \\
\midrule
Local branch only & 96.68 & 99.88 \\
Residual branch only & 6.52 & 88.00 \\
Local + Residual branches & 99.99 & 99.99 \\
\bottomrule
\end{tabular}
\caption{Ablation of branch components on \textsc{Boxes} at block size \(m{=}n//3\). Results are averaged over five random seeds (std.\ below $1.5\%$ for all entries).}
\label{tab:ablation_branches}
\end{table}

\subsection{Attention sparsity as a memory lever}\label{sec:sparsity-memory}

If attention mass concentrates on few keys, pruning each row to its top-$k$ entries (with renormalization) can cut memory with minimal loss. We study this under two protocols.

\textbf{Operators.}
We compare \(\mathcal{T}_{\mathrm{raw}}(A,V)=AV\), \(\mathcal{T}_{\mathrm{post}}(A,V)=A(I-A)^{-1}V\), and \(\mathcal{T}_{\mathrm{pre}}(A,V)=(I-A)^{-1}AV\).

\textbf{Top-$k$ construction.}
For row \(A_{i:}\), let \(\mathrm{TopK}_k(i)\) be indices of its \(k\) largest (causal) entries. The pruned, renormalized attention is
\[
\widehat{A}^{(k)}_{ij}=
\begin{cases}
\displaystyle \frac{A_{ij}}{\sum_{t\in \mathrm{TopK}_k(i)} A_{it}}, & j\in \mathrm{TopK}_k(i),\\[0.6em]
0, & \text{otherwise.}
\end{cases}
\]

\textbf{Protocol A: train \& test masking (robustness).}
Apply top-$k$ inside the target operator during training and testing.
Table~\ref{tab:op-topk} shows \(\mathcal{T}_{\mathrm{raw}}\) remains accurate even at \(k{=}1\), whereas inverse-involving operators degrade sharply when masked.

\begin{table}[t]
\centering
\begin{tabular}{l c}
\toprule
Operator (train \& test masking) & Accuracy \\
\midrule
$\mathcal{T}_{\text{raw}}(\widehat{A}^{(1)},V)$         & 100\% \\
$\mathcal{T}_{\text{post}}(\widehat{A}^{(1)},V)$        & 61\% \\
$\mathcal{T}_{\text{post}}(\widehat{A}^{(2)},V)$        & 88\% \\
$\mathcal{T}_{\text{post}}(\widehat{A}^{(3)},V)$        & 91\% \\
$\mathcal{T}_{\text{pre}}(\widehat{A}^{(1)},V)$         & 35\% \\
\bottomrule
\end{tabular}
\caption{Protocol A: enforcing top-$k$ at train \& test. Raw attention is robust; inverse-based operators are not. Results are averaged over three random seeds (std.\ below $2\%$ for all entries).}
\label{tab:op-topk}
\end{table}

\textbf{Protocol B: test-time pruning only (inherent sparsity).}
Train densely before replacing, at evaluation, \(A\) by \(\widehat{A}^{(k)}\) in \(\mathcal{T}_{\mathrm{raw}}\).
Accuracy/EM remain high for \(k\le3\) (Table~\ref{tab:boxes-topk}), indicating learned attention is naturally peaked.

\begin{table}[t]
\centering
\begin{tabular}{c c c}
\toprule
$k$ & $\mathcal{T}_{\text{raw}}(\widehat{A}^{(k)},V)$ Acc. & EM \\
\midrule
1 & 98.8\%  & 68.41\% \\
2 & 99.01\% & 75.12\% \\
3 & 99.89\% & 96.74\% \\
\bottomrule
\end{tabular}
\caption{Protocol B: train dense, prune only at test. High Acc/EM even for very small $k$.}
\label{tab:boxes-topk}
\end{table}

\textbf{Why pruning $A$ works (and pre-inverse pruning does not).}
Using \((I-A)^{-1}=\sum_{t\ge0}A^t\) gives \(A(I-A)^{-1}=\sum_{t\ge1}A^t\): one-hop routing (\(AV\)) hinges on the largest edges preserved by top-$k$, whereas multi-hop terms \(A^t\) densify; pruning \emph{before} the inverse removes low-weight bridges critical to long-range flow, hurting \(\mathcal{T}_{\mathrm{post}}\) and \(\mathcal{T}_{\mathrm{pre}}\).

\noindent\textbf{Memory note.}
Keeping $k$ entries per row reduces nonzeros from \(\Theta(hn^2)\) to \(\Theta(hnk)\); small \(k\in\{1,2,3\}\) therefore yields large storage savings while maintaining accuracy under Protocol B.

\subsection{Capacity limits: multi-property tracking}
\label{sec:capacity-multiprop}

\textbf{Setup.}
We extend \textsc{Toy} so that each entity carries $p$ independent properties, each evolving along its own chain (disjoint update rules and targets). The model reads a single sequence interleaving updates for all properties and must output the final value of \emph{each} property. We vary the number of attention heads $h$ and the number of properties $p$ while keeping the rest of the architecture fixed.

\textbf{Results.}
Table~\ref{tab:heads-props} shows a clear trend: with one head ($h{=}1$), performance collapses once $p{>}1$; increasing $h$ restores accuracy up to $h{=}p$ (e.g., $h{=}2$ succeeds for $p{=}2$, $h{=}4$ for $p{=}4$), and degrades again when $p$ outstrips $h$ (e.g., $p{=}8$ with $h{=}4$). The results obtained on \textsc{Boxes}, Tables~\ref{tab:boxes},~\ref{tab:op-topk},~\ref{tab:boxes-topk} (effectively a single property), $h{=}1$ attains the same accuracy as $h{=}8$ (100\%).

\begin{table}[t]
\centering
\small
\begin{tabular}{ccc}
\toprule
Heads $h$ & Properties $p$ & Accuracy \\
\midrule
1 & 1 & 100\% \\
1 & 2 & 61.8\% \\
2 & 2 & 100\% \\
4 & 4 & 100\% \\
4 & 8 & 61.3\% \\
\bottomrule
\end{tabular}
\caption{Heads vs.\ properties on multi-property \textsc{Toy}. Accuracy is high when $h\!\ge\!p$ and drops when $p$ exceeds $h$. Results are averaged over three random seeds (std.\ below $0.5\%$ for all entries).}
\label{tab:heads-props}
\end{table}

\noindent\textbf{Interpretation, implications, and guidance.}

Let $h$ be the number of attention heads and $p$ the number of simultaneously evolving properties. Empirically, accuracy is high when $h\!\ge\!p$ and degrades when $p\!>\!h$ (Table~\ref{tab:heads-props}), consistent with \emph{interference}: when several properties share a head, their routing signals superpose and become corrupted. 

Increasing $h$ provides separate routing channels, explaining the sharp transitions in Table~\ref{tab:heads-props} and the parity between $h{=}1$ and $h{=}8$ on \textsc{Boxes} (effectively $p{\approx}1$). The same constraint applies under our blockwise evaluation: blockwise computation changes \emph{efficiency}, not the representational requirement that $h$ scale with the number of concurrent properties. In practice, one head suffices for single-attribute tracking; for multi-attribute tracking, use $h\!\approx\!p$ (or sweep $h$ until performance saturates).

\subsection{Language Modeling Capabilities}\label{sec:lm}

We evaluate the language modeling capability of our blockwise ChaCAL attention within a GPT-2 architecture (\(L{=}12\), \(h{=}12\), \(d_{\mathrm{model}}{=}768\), \(d_{\mathrm{ff}}{=}3072\)) using context length \(n=1024\) unless stated otherwise.
We report two regimes in a unified format: (i) fine-tuning from a GPT-2 model pretrained with dense attention on OpenWebText, and (ii) training from scratch on WikiText-103 to remove the pretraining-mismatch confound.
In all cases, we replace attention in all 12 layers following the replacement protocol (Section~\ref{sec:models}) and sweep block sizes \(m\) for our blockwise operator.

\begin{table}[t]
\centering
\small
\resizebox{\columnwidth}{!}{
\begin{tabular}{@{}l l r@{}}
\toprule
\textbf{Setting} & \textbf{Model} & \textbf{PPL} \\
\midrule
\multicolumn{3}{l}{\textbf{OpenWebText (fine-tune from pretrained GPT-2)}}\\
\midrule
Pretrained & GPT-2 (pre-trained) & 23.78 \\
Fine-tuned & GPT-2 (dense attention) & 20.15 \\
Fine-tuned & GPT-2 + ChaCAL & 21.46 \\
Fine-tuned & GPT-2 + Block-ChaCAL (ours, $m{=}n//2$) & 22.60 \\
Fine-tuned & GPT-2 + Block-ChaCAL (ours, $m{=}n//3$) & 22.08 \\
Fine-tuned & GPT-2 + Block-ChaCAL (ours, $m{=}n//4$) & 22.60 \\
\midrule
\multicolumn{3}{l}{\textbf{WikiText-103 (train from scratch)}}\\
\midrule
From scratch & GPT-2 (dense attention) & 16.46 \\
From scratch & GPT-2 + ChaCAL & 17.57 \\
From scratch & GPT-2 + Block-ChaCAL (ours, $m{=}n//2$) & 15.96 \\
From scratch & GPT-2 + Block-ChaCAL (ours, $m{=}n//3$) & 15.64 \\
From scratch & GPT-2 + Block-ChaCAL (ours, $m{=}n//4$) & 15.18 \\
\bottomrule
\end{tabular}}
\caption{Language modeling perplexity (PPL; lower is better) for GPT-2 variants under two settings: OpenWebText fine-tuning from a pretrained GPT-2, and WikiText-103 training from scratch. “Ours” denotes the proposed blockwise resolvent operator.}
\label{tab:lm_all}
\end{table}

Table~\ref{tab:lm_all} highlights two complementary observations.
First, under OpenWebText fine-tuning (pretraining-matched distribution), dense GPT-2 retains the best PPL, while ChaCAL and blockwise variants are slightly worse, consistent with architectural mismatch relative to the pretrained weights.
Second, when trained from scratch on WikiText-103, Block-ChaCAL outperforms both dense attention and ChaCAL, showing that the blockwise operator is not intrinsically weaker than the baseline and can provide a favorable inductive bias in language modeling.

\textbf{Train/test context length dependence (WikiText-103, Block-ChaCAL, $m{=}n//3$).}
We further evaluate how performance changes when training and testing at different context lengths.
This experiment probes length extrapolation and distribution shift in \(n\) (Table~\ref{tab:lm_length_sweep}); showing performance degradation under strong length extrapolation.

\begin{table}[t]
\centering
\small
\begin{tabular}{c c c}
\toprule
\textbf{Train }$n$ & \textbf{Test }$n$ & \textbf{Valid PPL} \\
\midrule
1024 & 256 & 21.98 \\
1024 & 512 & 17.99 \\
1024 & 1024 & 15.64 \\
256 & 256 & 13.33 \\
256 & 512 & 28.79 \\
256 & 1024 & 51.42 \\
\bottomrule
\end{tabular}
\caption{WikiText-103: dependence on train/test context length for Block-ChaCAL at block size $m{=}n//3$. Performance degrades under strong length extrapolation.}
\label{tab:lm_length_sweep}
\end{table}

\subsection{Code reasoning as natural entity tracking}\label{sec:code-entity-tracking}
\begin{table}[ht]
\centering
\small
\setlength{\tabcolsep}{3pt}
\begin{tabular}{@{}l r@{}}
\toprule
\textbf{Model} & \textbf{Accuracy} \\
\midrule
Transformer (fine-tuned) & 82.40\% \\
ChaCAL (fine-tuned) & 85.36\% \\
ChaCAL (block, \(m{=}n//4\), fine-tuned) (ours) & 83.67\% \\
ChaCAL (block, \(m{=}n//3\), fine-tuned) (ours) & 83.65\% \\
\bottomrule
\end{tabular}
\caption{Fine-tuning results on the \textit{Copilot} dataset under the controlled substitution protocol (only layer 2).}
\label{tab:copilot_results}
\end{table}

Code understanding inherently requires tracking entities—variables, objects, and functions—through updates, scopes, and call chains. This mirrors our \textsc{Boxes}/\textsc{Toy} setup: tokens act as "state carriers" whose values must be routed and updated across long ranges. We therefore test whether mechanisms that excel at entity tracking also help on code.
We use a controlled substitution design to isolate the effect of the attention operator while minimizing distribution shift from the pretrained backbone: only one layer is modified, and all other weights remain in-distribution.
We choose layer 2 as an early routing layer that can influence variable binding and entity propagation without fully disrupting the model’s positional processing in the first layer.

Table~\ref{tab:copilot_results} shows that ChaCAL achieves the highest accuracy (85.36\%), surpassing the Transformer baseline (82.40\%). Our blockwise variants reach 83.67\% (\(m{=}n//4\)) and 83.65\% (\(m{=}n//3\)), both above the baseline, indicating that code tasks can benefit from explicit entity-tracking operators.

\subsection{Encoder-decoder transfer: long-range SCROLLS with BART}
\label{sec:scrolls_bart}
\begin{table}[t]
\centering
\resizebox{\columnwidth}{!}{
\small
\setlength{\tabcolsep}{5pt}
\begin{tabular}{@{}l l r r r@{}}
\toprule
\textbf{Task} & \textbf{Metric} 
& \textbf{zero-shot} 
& \textbf{FT (standard)} 
& \textbf{FT (modified)} \\
\midrule

\multirow{4}{*}{\textsc{GovReport}} 
& ROUGE-1   & 29.68 & 48.18 & 38.46 \\
& ROUGE-2   & 10.84 & 23.55 & 11.52 \\
& ROUGE-L   & 16.32 & 29.26 & 20.12 \\
& ROUGE-GM  & 16.99 & 31.64 & 20.43 \\
\midrule

\multirow{1}{*}{\textsc{NarrativeQA}} 
& F1        & 2.39 & 11.63 & 7.86 \\
\midrule

\multirow{1}{*}{\textsc{ContractNLI}} 
& EM        & 30.63 & 59.38 & 36.72 \\
\bottomrule
\end{tabular}
}
\caption{SCROLLS results for BART-base at input length up to \(n{=}1024\). Zero-shot refers to pretrained BART evaluated without fine-tuning. “FT (standard)” is standard fine-tuning with dense attention. “FT (modified)” fine-tunes after substituting the mechanism in the decoder's first layer, with $m=n//3$.}
\label{tab:scrolls}
\end{table}

While the above experiments focus on fully causal backbones (GPT-2 style), many long-context downstream benchmarks are commonly approached with \emph{encoder-decoder} models such as BART~\citep{lewis2020bart}.
We therefore probe how Block-ChaCAL behaves under this architectural regime, using SCROLLS~\citep{shaham2022scrolls} tasks as a stress test for long-range understanding and report the results in Table~\ref{tab:scrolls}.

Dense-attention BART fine-tuning remains best, while swapping in a single Block-ChaCAL decoder layer typically degrades performance relative to that dense fine-tuned baseline.
However, our proposition still improves substantially over BART zero-shot, indicating the model can adapt but is less plug-and-play under the same recipe.
This differs from fully causal decoder-only settings (Section~\ref{sec:code-entity-tracking}), where a controlled single-layer substitution can outperform dense attention.

In encoder--decoder models, pretrained bidirectional encoder representations and the decoder’s attention dynamics are tightly coupled under dense attention; changing one decoder layer alters routing in a way the pretrained weights were not optimized for.
Two practical directions to modify this behavior follow: (i) an adaptation stage before supervised fine-tuning, or (ii) focusing on fully causal architectures where the operator better matches the masking and objective.

\section{Discussion and Conclusion}
\label{sec:conclusion}
We presented a blockwise evaluation of a resolvent-style, entity-tracking attention layer that is exact on local tiles and routes cross-block interactions through a compact reduced system, yielding subquadratic sequence cost \(\widetilde{\mathcal{O}}(n^{4/3}d)\) (or \(\widetilde{\mathcal{O}}(n^{7/3})\) when \(d{\approx}n\)) while preserving the semantics of the dense operator of~\citet{Fagnou_2024}. 
In practice, our method is most effective for causal, entity-centric tasks with mostly local routing and enough heads to track concurrent properties. In this regime, Block-ChaCAL retains the benefits of resolvent-style propagation while lowering evaluation cost. When routing is diffuse or the model is strongly tied to dense-attention pretraining, gains are less immediate and may require adaptation.

On controlled tracking tasks, it matches dense accuracy and yields consistent wall-clock gains, reaching the EM of a 5-layer dense Transformer at \(\sim2.4{\times}\) lower latency. In fully causal language modeling from scratch, Block-ChaCAL matches or improves over dense attention, while fine-tuning from pretrained weights incurs a small mismatch cost. On SCROLLS with BART-base, the Block-ChaCAL variant remains above zero-shot but typically trails standard dense fine-tuning, suggesting encoder--decoder adaptation is less plug-and-play under an unchanged recipe. 

Overall, results support a structured-sparsity view: attention concentrates within local neighborhoods with a light off-block residue, which our exact-on-blocks plus reduced residual path captures efficiently; Top-\(k\) diagnostics likewise indicate one-hop routing is highly compressible, whereas inverse-involving forms densify and should be handled in compressed space.
A capacity constraint remains: accuracy is high when \(h\!\ge\!p\) and drops when \(p\!>\!h\); efficiency gains do not remove representational needs. In practice, one head suffices for single-attribute tracking and \(h\!\approx\!p\) is beneficial for multi-attribute tracking; block sizes \(n//2\)–\(n//6\) offer strong latency--quality trade-offs near the runtime minimum.

Promising directions include staged adaptation to reduce mismatch, adaptive block sizing and learned down/up-sampling, head-aware residual routing, and kernel-level optimization. Overall, for entity-centric reasoning, attention is structured, and exploiting this structure yields practical speedups while preserving resolvent behavior.




\newpage
\section*{Limitations}

Our approach assumes \emph{block-structured sparsity}: most attention mass remains within local tiles with a light, structured off-block residue. Tasks with diffuse, global dependencies (e.g. multi-document aggregation) may exhibit weaker speedups. Efficiency gains hinge on a block size $m$ near the empirical balance point; far from this regime, the reduced-system cost or local-tile cost can dominate, yielding a U-shaped runtime curve. Although the reduced cross-block computation is small in our settings, its $k{\times}k$ solve (\(k{=}n/m\)) adds overhead that can become non-negligible if $m$ is chosen too small. Representational capacity is still constrained by the \emph{head–property} relationship: when the number of simultaneously evolving properties $p$ exceeds heads $h$, accuracy degrades; blockwise evaluation does not remove this requirement. 

A further limitation is \emph{pretraining mismatch}. Substituting the attention operator inside a pretrained model changes the computation graph seen during pretraining. In our fine-tuning experiments, this mismatch is mitigated for decoder-only models by modifying a single layer; however, for encoder-decoder architectures the adaptation can be less straightforward. In particular, we observed that a standard fine-tuned encoder-decoder model can outperform its variant under the same fine-tuning recipe. This suggests that realizing the benefits of the operator in pretrained encoder-decoder models may require (i) pretraining with the modified attention before fine-tuning, or (ii) focusing on fully causal backbones where the operator can be introduced locally and the controlled substitution protocol is better aligned with the pretrained distribution.

While we report language modeling parity with dense attention/dense resolvent in our experiments, broader pretraining/evaluation (domains, languages, noise conditions) is needed to fully assess generalization. Reported wall-clock gains are hardware- and implementation-dependent; different kernels, memory bandwidth, and batching can shift absolute timings, even if asymptotics remain. 
Accordingly, implementing a dedicated CUDA kernel for our method is the natural next step to enable a fair comparison with FlashAttention variants.
Finally, our analysis and experiments focus on causal/structured masks; applicability to non-causal or highly irregular masking patterns may require additional engineering (e.g., mask-aware down/up-sampling) and could reduce savings.

\section*{Acknowledgments}
This work was granted access to the HPC resources of IDRIS under the allocations A0191016927 and AD011015154R2 made by GENCI. This work has received support from the French government, managed by the National Research Agency, under the France 2030 program with the reference “PR[AI]RIE-PSAI” (ANR-23-IACL-0008) and "PEPR-SHARP" (ANR-23-PEIA-0008).

\bibliography{bibliography}
\newpage
\appendix
\section{Extended Related work}
\label{app:extended_related}
\subsection{Self-attention mechanism}
The attention mechanism \citep{NIPS2017_3f5ee243} lies at the core of the Transformer, enabling long-range dependencies by computing interactions between all token pairs. Given $X\!\in\!\mathbb{R}^{n\times d}$,
\begin{equation}
\begin{aligned}
    &A \,=\, \mathrm{Softmax}\!\left(\frac{(XW_q)(XW_k)^{\top}}{\sqrt{d_k}} + M \right), \\
    &Y \,=\, A X W_v \,=\, A V,
\end{aligned}
\end{equation}
with learnable $W_q\!\in\! \mathbb{R}^{d\times d_k},\,W_k\!\in\!\mathbb{R}^{d\times d_k},\,W_v\!\in\!\mathbb{R}^{d\times d_v}$ and mask $M$ (causal or structural). While expressive, dense self-attention has at least quadratic cost in $n$, motivating structure-aware operators \citep{kim2017structured} and a large body of efficient variants reviewed below.

\subsection{Entity tracking}
Entity tracking maintains and updates entity-centric state across discourse. Early work modeled local coherence with entity grids \citep{barzilay2008entity} and leveraged entity-centric coreference \citep{clark2016entity}. Memory architectures such as Recurrent Entity Networks explicitly maintain entity states \citep{henaff2017tracking}. With Transformers, end-to-end coreference became a strong neural baseline \citep{lee2017end}, and analyses showed attention often focuses on entity-relevant patterns \citep{clark2019entity}. Task-tailored layers make this explicit: Entity-Aware Transformers inject entity representations into attention \citep{gerritse2022entity}, and Chain and Causal Attention (ChaCAL) compresses multi-layer propagation for entity tracking into a single operator \citep{Fagnou_2024}. Beyond datasets, recent probes examine how LMs handle state updates \citep{kim-schuster-2023-entity} and how fine-tuning surfaces latent mechanisms \citep{prakash2024finetuningenhancesexistingmechanisms}.
A key theoretical point for depth is that standard stacked attention must grow with the \emph{number of state changes} $K$ to propagate information: \(\mathcal{O}(\log K)\) layers are sufficient (and in some cases necessary) for $K$ updates \citep{Fagnou_2024}. This motivates operators that \emph{compress depth} while preserving entity-centric semantics, which is precisely the angle we pursue with our blockwise inverse formulation.

\subsection{Efficient attention and long-context modeling}
To reduce the compute/memory burden, many works introduce structure into attention: block/strided sparsity \citep{child2019sparse}, LSH-based nearest-neighbor routing \citep{kitaev2020reformer}, sliding windows with global tokens \citep{beltagy2020longformer}, block-sparse plus random links with guarantees \citep{zaheer2020bigbird}, low-rank projections \citep{wang2020linformer}, and kernelized softmax via random features \citep{choromanski2021performer}. Standardized long-range benchmarks such as LRA \citep{tay2021long} and SCROLLS \citep{shaham2022scrolls} emphasize both length generalization and reasoning. Orthogonally, FlashAttention makes exact attention IO-aware~\citep{dao2022flashattentionfastmemoryefficientexact}, and recent analyses suggest many attention connections are redundant in practice \citep{he2024matterstransformersattentionneeded}. Beyond attention patterns, token-efficiency approaches (dynamic pooling/pruning and length adaptation) cut cost by reducing effective sequence length with minimal quality loss \citep{nawrot2023efficient,rao2021dynamicvit,kim2022lengthadaptive,zeng2022not}, and hierarchical merging explores semantic condensation for long contexts \citep{song2024hierarchical}.

Our work is complementary: instead of approximating generic attention or only optimizing kernels, we exploit the \emph{intrinsic, structured sparsity} that emerges in entity tracking to design a \emph{task-specific} attention operator with a fast path and \emph{subquadratic} sequence complexity, while retaining the semantics that matter for entity-centric reasoning \citep{kim2017structured,gerritse2022entity,Fagnou_2024}.

\section{Complexity Analysis Details}
\label{sec:appendix_complexity_details}

\textbf{Load per branch.}
Let $n=km$ and treat width $d$ as independent of $n$.
\begin{itemize}[nosep]
\item Local branch:
\begin{align*}
    &\quad \sum_{i=1}^k \mathcal{O}(m^2 d) \;=\; \mathcal{O}(k\,m^2 d)\;=\;\mathcal{O}(n m d).
\end{align*}
\item Residual down/up:
\begin{align*}
&\quad \mathcal{O}(nd)\ \text{(negligible vs.\ leading terms).}
\end{align*}
\item Reduced operator:
\begin{align*}
\quad \mathcal{O}(k^2 d)\ &\text{for forming products}
\\
&+\;\text{solve on }k{\times}k.
\end{align*}
\end{itemize}

\textbf{Balancing and overall complexity.}
The dominant terms are $\mathcal{O}(n m d)$ (local) and $\mathcal{O}((n/m)^2 d)$ (reduced). Minimizing
\begin{equation}
\begin{aligned}
T(n,d;m) \;&=\; c_1\, n m d \;+\; c_2\, (n/m)^2 d
\end{aligned}
\end{equation}
w.r.t.\ $m$ yields $m^\star$ asymptotically equal to $n^{2/3}$ and
\begin{equation}
\begin{aligned}
T(n,d;m^\star) \;&=\; \widetilde{\mathcal{O}}(n^{4/3} d).
\end{aligned}
\end{equation}
When $d$ scales with $n$ (e.g., $d\!\approx\!n$), the overall cost becomes $\widetilde{\mathcal{O}}(n^{7/3})$. Down/up-sampling $\mathcal{O}(nd)$ and any $k^3$ dense factorization (if used instead of \texttt{triSolve}) are dominated in the balanced regime since $k=n/m=\Theta(n^{1/3})$.

\section{Detailed Experimental Settings}
\label{app:exp_details}

This appendix specifies the concrete data, model, and training configurations used throughout the paper. Unless explicitly stated, all compared methods share the same preprocessing, optimizer, batch sizing, and decoding settings; the only change is the attention operator and the corresponding replacement scope. The code will be open sourced upon publication.

\subsection{Tasks, datasets, and metrics}
\label{app:datasets_metrics}

We evaluate on four families of tasks.

\textbf{Toy (Chains).}
Synthetic multi-hop tracking with fixed length \(n{=}128\) (16 blocks of 8 tokens) and chain depth up to 15 (Section~\ref{sec:tasks-metrics}). Metric: token accuracy. When used, Top-\(k\) masking is applied row-wise to the (masked) attention probabilities with renormalization.

\textbf{Boxes.}
The advanced \textsc{Boxes} entity-tracking task of \citet{Fagnou_2024}. Each instance describes an initial state and a sequence of Put/Remove/Move operations (including implicit ``move the contents'' operations). Metrics: token accuracy and exact match (EM) after normalization.

\textbf{Language modeling and code reasoning.}
We report perplexity (PPL) on language modeling at context length \(n{=}1024\), and classification accuracy on the code reasoning setup described in Section~\ref{sec:code-entity-tracking}.

\textbf{SCROLLS (long-context seq2seq).}
We evaluate \texttt{BART-base} on SCROLLS tasks spanning summarization (ROUGE), QA (token-level F1), and classification-style tasks (EM), with maximum input length \(n\le1024\) and maximum target length 256. For SCROLLS, we select the best validation checkpoint by the task’s primary metric (ROUGE geometric mean / F1 / EM).

\subsection{Models and replacement protocols}
\label{app:models_protocols}

\textbf{Operators.}
We compare dense attention, ChaCAL \citep{Fagnou_2024}, and our proposition, which evaluates the resolvent exactly on block-diagonal \(m{\times}m\) tiles and handles cross-tile interactions via a reduced \(k{\times}k\) system with \(k{=}n/m\). We sweep \(m\in\{n//2,n//3,\ldots,n//8\}\) unless stated otherwise; \(m{=}n\) recovers ChaCAL.

\textbf{Controlled replacement (default).}
To isolate mechanism effects while limiting distribution shift, we modify only the layers specified below and keep all other components unchanged:
\emph{Toy}: replace the sole attention layer.
\emph{Boxes}: replace only the mechanism layer (the other layer remains standard).
\emph{LM}: replace attention in all 12 layers.
\emph{Code}: replace only layer 2 (controlled substitution).

\textbf{BART protocol for SCROLLS.}
We keep the \emph{encoder non-causal} (standard bidirectional self-attention) and replace \emph{only the first decoder self-attention layer} by our mechanism; all encoder layers, decoder cross-attention, and remaining decoder self-attention layers stay dense. This mirrors the controlled-substitution design: an early decoder routing layer can influence long-range propagation during generation while preserving most pretrained structure.

\subsection{Training, checkpointing, and decoding}
\label{app:training_decoding}

\textbf{Non-SCROLLS experiments.}
Unless otherwise stated in the main text, we train with AdamW (learning rate \(3\times10^{-4}\)), batch size 256, and up to 25k steps on a single NVIDIA A6000. Timings are reported as average evaluation time over the full evaluation set with identical batch size across methods.

\textbf{SCROLLS (BART-base).}
We fine-tune \texttt{facebook/bart-base} with AdamW (learning rate \(5\times10^{-5}\), warmup ratio 0.1), effective batch size 8 (per-device 1, gradient accumulation 8), for 1 epoch. We evaluate and save every 500 steps, use greedy decoding (\(\texttt{num\_beams}=1\)), and cap evaluation generation length to the desired long-range diagnostic value. For compatibility with attention modifications, we set \texttt{use\_cache=False}; mixed precision and checkpointing are enabled only when needed for memory and are recorded per run.

\subsection{Logging and model selection}
\label{app:repro_logging}

For each run, we store the full configuration (hyperparameters, block size, and replacement scope) alongside checkpoints, and report the best validation score achieved across saved checkpoints according to the task’s primary metric. When multiple seeds are used, we report mean results; variability is typically small in the controlled synthetic settings.



\end{document}